\documentclass[10pt,twocolumn,letterpaper]{article}
\usepackage{cvpr}
\usepackage{times}
\usepackage{epsfig}
\usepackage{graphicx}
\usepackage{amsmath}
\usepackage{amssymb}
\usepackage{csquotes}
\graphicspath{{./fig/}}

\usepackage[breaklinks=true,bookmarks=false]{hyperref}
\cvprfinalcopy 
\def\cvprPaperID{3499} % *** Enter the CVPR Paper ID here

\begin{document}

%%%%%%%%% TITLE
\title{ Privacy-Protective-GAN for Face De-identification}

\author{Yifan Wu, Fan Yang, Haibin Ling\\Temple University \\\tt\small  \{yifan.wu, fyang, hbling\} @temple.edu}

\maketitle
%\thispagestyle{empty}

%%%%%%%%% ABSTRACT
\begin{abstract}
Face de-identification has become increasingly important as the image sources are explosively growing and easily accessible. The advance of new face recognition techniques also arises people’s concern regarding the privacy leakage. The mainstream pipelines of face de-identification are mostly based on the k-same framework, which bears critiques of low effectiveness and poor visual quality. In this paper, we propose a new framework called Privacy-Protective-GAN (PP-GAN) that adapts GAN with novel verificator and regulator modules specially designed for the face de-identification problem to ensure generating de-identified output with retained structure similarity according to a single input. We evaluate the proposed approach in terms of privacy protection, utility preservation, and structure similarity. Our approach not only outperforms existing face de-identification techniques but also provides a practical framework of adapting GAN with priors of domain knowledge.
\end{abstract}
%-------------------------------------------------------------------------
\section{Introduction}
Beneficial from the blooming development of media and network techniques that makes huge amount of images more approachable,  image analysis techniques bear its prosperity in the past decade and brings unprecedented convenience to our daily life. Among those image sources exposed to the public with or without our awareness, a considerable number of them contain our  identity especially the biometric information. Not only will the unprotected exposing cause the leak of privacy, the common approaches of protection like blurring and pixelization may also not be satisfied in thwarting face recognition software \cite{newton2005preserving,ribaric2016identification}. The other extreme side is that we simply mask identity area off, which is perfect for identity removal. But it causes serious loss of data utility in application of visual understanding as the scene information is changed with objects removal. Thus it is critically important to build a framework that can properly de-identify the privacy information from the image while keeping its utility at the same time.

Specially for the face de-identification problem, the dilemma is that on the one hand,  we want the de-identified image to look as different as possible from the original image to ensure the removal of identity; on the other hand, we expect the de-identified image to retain as much structural information in the original image as possible so that the image utility remains. Previous approaches on this problem are mostly based on k-same algorithm \cite{gross2005integrating,  gross2006model, gross2008semi, newton2005preserving} for the de-identificatin procedure and some apply additional models like the Active Appearance Models (AAMs) \cite{matthews2004active} to explicitly construct faces preserving the utility attributes like gender, race, and age \cite{du2014garp, jourabloo2015attribute}. Yet these models fail to make full use of existing data and deliver fairly poor visual quality due to the unnatural synthesis.

The Generative Adversarial Networks (GANs) provide an inspiring framework on generating sharp and realistic natural image samples via adversarial training \cite{goodfellow2014generative,isola2016image, zhu2017unpaired}. GAN can be naturally used for the face de-identification as it can generate new samples from the gallery following original input data distribution. Since GAN implicitly models a dataset distribution, it cannot involve processing individual images. To achieve an image-to-image face de-identification, conditinal GAN (cGAN) is intuitively leveraged here, which is an extention of GAN that fits a conditional data distribution \cite{cGAN2014conditional}.

\begin{figure}[t]	
	\begin{center}
		%\includegraphics[width=0.95\linewidth, height=0.3\textheight,keepaspectratio]{nima4.png}
		\includegraphics[width=\linewidth,keepaspectratio]{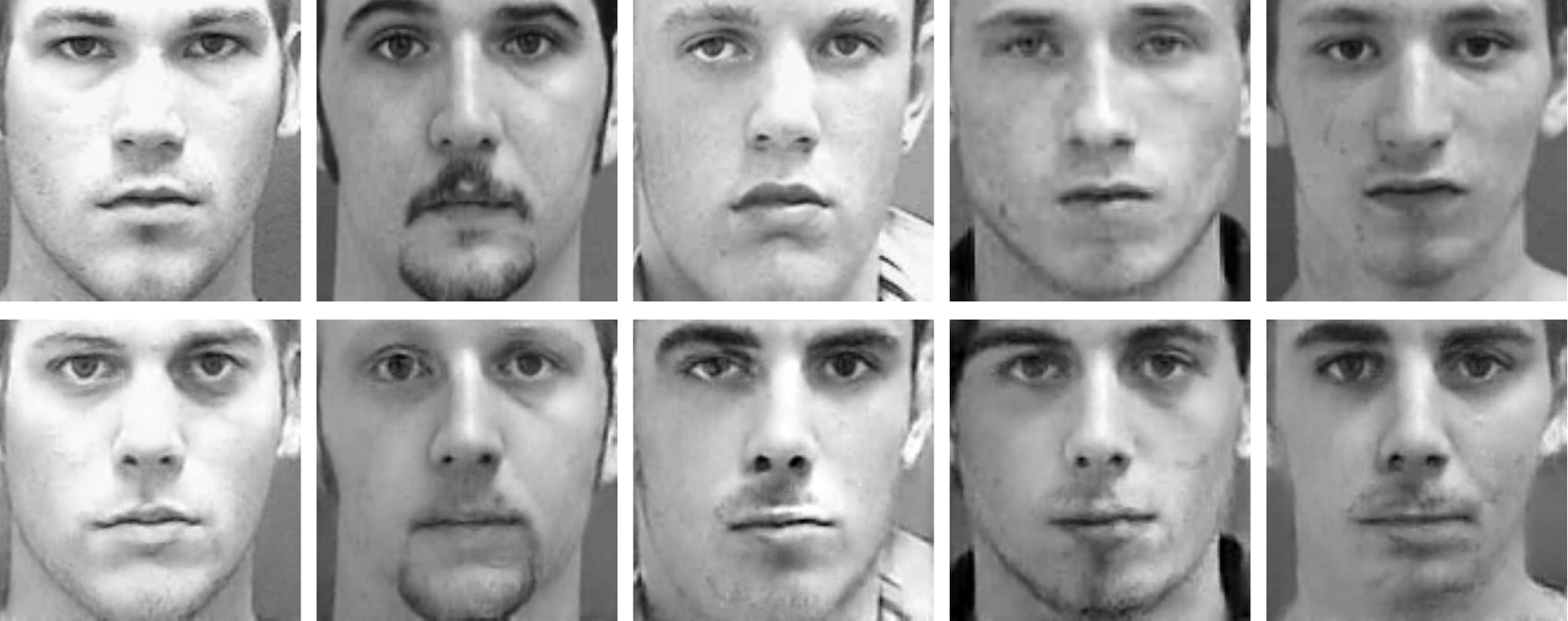}
	\end{center}
	\caption{
		\textbf{Illustration of face de-identification.} The top row shows the original face images, the bottom row represents corresponding de-identified images synthesized by our method.}
	\label{fig:demo}
\end{figure}
%The extention of the original GAN from following full data distribution to conditional data distribution is called conditinal GAN (cGAN), the generator of which can be naturally used for the de-identification here as it can generate a new sample from the gallery following the same conditional distribution with the original input image, where the sampling procedure should be able to realize the id-removal and the conditional setting may probably help keep the same structural information as the original one.
Unfortunately, cGAN is not directly applicable to our scenario, which requires privacy protection as well as data utility preservation. First, the generative loss in cGAN is not specific for distinguishing identities, thus the generated samples could be either too similar with each other to pass external face verification, or visually ghosted as none-faces. Second, the structure information is unensured to be preserved due to the inconsistency in feature spaces between GAN-type and structural losses.
%However, the cGAN are still not satisfying in two aspects.

\begin{figure*}[t]	
	\begin{center}
		%\includegraphics[width=0.95\linewidth, height=0.3\textheight,keepaspectratio]{nima4.png}
		\includegraphics[width=0.85\linewidth,keepaspectratio]{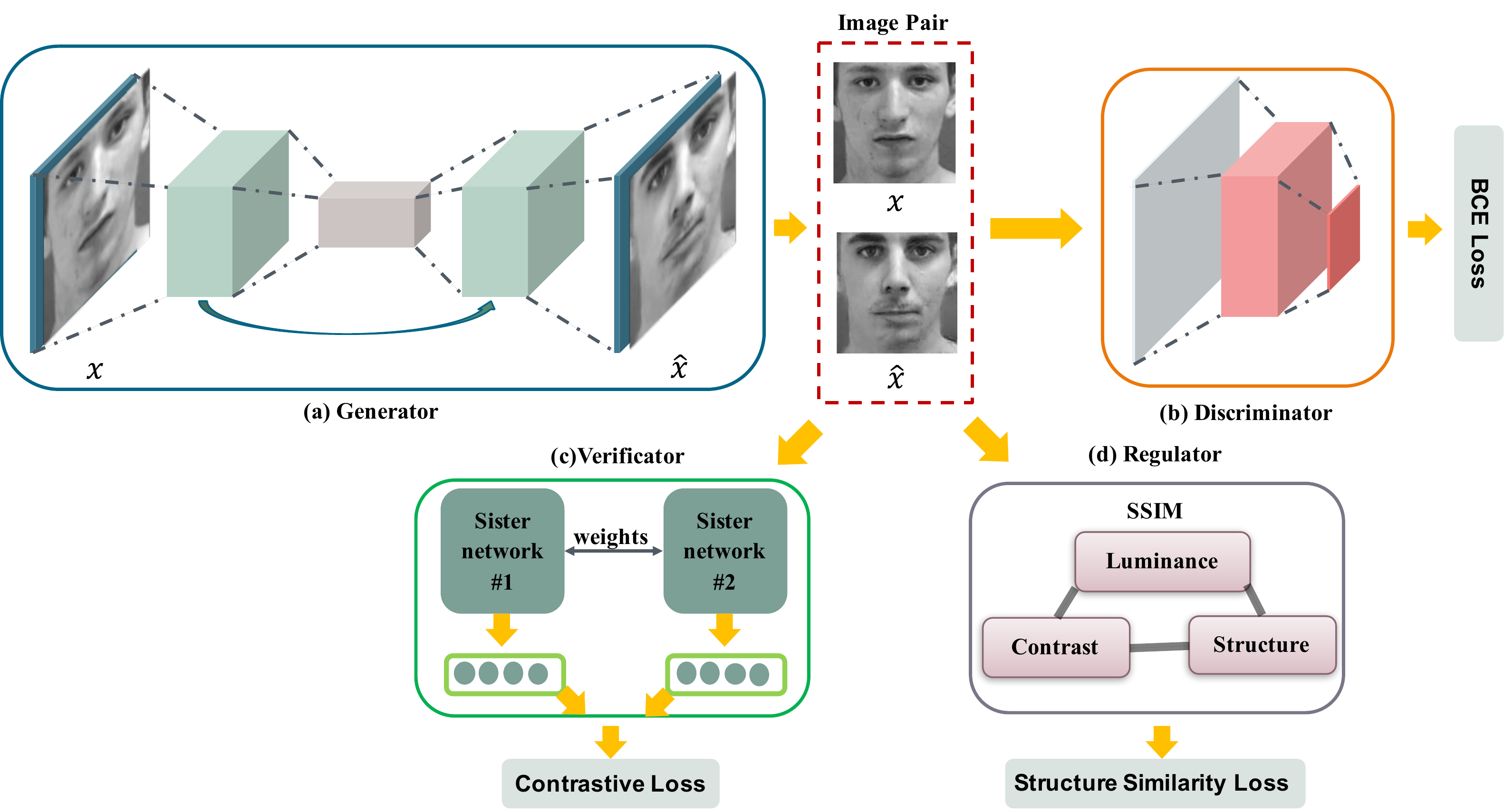}
	\end{center}
	\caption{
		\textbf{The Structure of the Privacy-Protective-GAN (PP-GAN) Framework.} The PP-GAN framework consists of four components: generator, discriminator, verificator, and regulator. \emph{(a)} The generator is a
		\enquote{U-Net} based auto-encoder, synthesizing a de-identified image $\hat{x}$ for given original image $\hat{x}$. \emph{(b)} The discriminator is adversarially trained with the generator to encourage the output to be sharp and realistic. \emph{(c)} The verifiactor is to determine whether two faces are from the same person. \emph{(d)} The regulator is to determine how similar the two faces are in pixel level.}
	\label{fig:framework}
\end{figure*}

Thus to make an adaptive GAN framework for the de-identification problem, we introduce additional mechanisms to enlarge the sampling range away from the given samples in the identity-related embedding feature space properly and to constrain the variation of the part of image structure from the full image feature space simultaneously. When unified in the GAN loss, these two tasks are actually contradictory to each other because the GAN loss is unable to explicitly tell the embedding feature from the structure feature and they contribute to the performance of distinguishing two faces in the same direction. Thus here we explicitly express them as external losses to compensate their contradiction in GAN and consequently achieve a novel Privacy-Protective-GAN model (PP-GAN, see Figure.~\ref{fig:framework}), which can plausibly balance between id-removal and structure retains.

 %Here we exploit this advance to develop an end-to-end network that synthesizes the de-identified images given original face inputs. We introduce a new counterwork to retain the structure similarity while holding enough distance in the identity-related feature space between input and output by adding a regulating item.
Indeed, in our PP-GAN model, we introduce two additional types of external modules: the verificator with contrastive loss \cite{hadsell2006dimensionality} to allow cGAN's generator for a larger range of sampling exploration, and the regulator with Structural Similarity Index (SSIM) \cite{wang2004image} to account for the structure preservation. Intuitively, the verificator adds prior information of identity feature matching in the embedding space to help remove biometic information while the regulator tells the generator how to maintain similar image utility via luminance, contrast and structural similarities. The involvement of these two prior knowledge on what is identity information versus the structure information significantly improve the model performance.

In the experimental setting, we quantitatively demonstrate the effectiveness of proposed method in several aspects: (\romannumeral1) The generative de-identified image can thwart the face verificator; (\romannumeral2) The de-identified image is not simply switched with others in the training data; (\romannumeral3) The luminance, contrast and structure can be partially preserved and (\romannumeral4) The attributes utility can be preserved when training an attribute specific generator.

In summary, our contributions are as following:

$\bullet$ To the best of our knowledge, we are the first to propose a GAN-based framework that is trainable in an end-to-end manner directly for the face de-identification task. The integrated framework can synthesize de-identified output for each single probe face.

$\bullet$ We manage to balance the trade-off between image quality and privacy protection by introducing novel modules, the verificator and the reglulator. The model learns to retain the structure similarity with the original image on the pix level, while distinguishes the input and synthesized output in the identity-related feature space. It is the first time that a de-identification metric is properly aggregated into the objective function in a controllable and measurable manner, which largely ensured the effectiveness of de-identification and privacy protection.

$\bullet$ We are the first to propose a systematic plan of evaluating the results of de-identification, where we employ a state-of-the-art face verificator \cite{schroff2015facenet} to determine the de-identification rate and the a face detector \cite{mtcnn} to quantify the detection rate. These two measurements combinely indicate how well the de-identification procedure works in its target of with utility retained. The proposed method achieves 100\% de-identification rate and obtains best trade-off between identity removal and visual similarity.

The rest of the paper is orgnaized as following: Related works about de-identification task are summarized in Section \ref{sec:related work}; In the Section \ref{sec:Method}, we introduce our PP-GAN framework; A quantitative evaluation on a public dataset as well as visual illustrations are provided in Section \ref{sec:Experiments}; The paper concludes with a discussion in Section \ref{sec:conlcusion}.

\section{Related Work}
\label{sec:related work}
%The problem of preserving privacy in no-visual data has been intensively studied and a thorough survey is given in \cite{FungWCY10survey}.
Privacy protection in visual data has been drawing increasing amount of research attention recently, with focuses on different scenarios such as video analysis and security and sampled studies in~\cite{Senior09ppvs,ChenCYY07jasp,Wilber&Boult12wacv,ChanLV08cvpr,DeCann&Ross12cvprw}

Face de-identification is an important tool for visual privacy protection. The goal of face de-identification has two folds: privacy protection and data utility preservation. In such way, de-identified images conceal the identity privacy of the original image, while preserving non-identity-related aspects for further data analysis. Earlier works on face de-identification simply use masking, blurring or pixelation \cite{ribaric2016identification}. While these Ad Hoc methods are easily applicable, it is shown that they provide no privacy assurance since they may fail to thwart face recognition software \cite{gross2009face}. Moreover, how to conduct a \emph{sufficient blur} itself is non-trivial~\cite{frome2009large}.

%A person de-identification method is proposed in~\cite{Agrawal&Narayanan11tcsvt} to de-identify a person but retain his/her action information. It implicitly use the human action as the data utility. However, it is a very modest data utility and many important attributes (e.g., gender) are lost.

To address this problem, Newton \etal \cite{newton2005preserving} propose the \emph{k}-same algorithm based on the \emph{k}-anonymity concept \cite{sweeney2002k}. By applying the \emph{k}-same algorithm, a given image is represented by average face of \emph{k}-closet faces from the gallery. This procedure theoretically limits the performance of recognition to $1/k$, but usually suffers from ghosting artifacts in de-identified images. Some variants of \emph{k}-same are proposed to improve the data utility and the naturalness of de-identified face images.
The \emph{k}-Same-Select algorithm \cite{gross2005integrating} partitions the image set into mutually exclusive subsets and applies the \emph{k}-same algorithm independently to each subsets, in this way attempts to preserve attributes of each subset.
In order to overcome undesirable artifacts due to misalignments and produce de-identified images of better quality, the \emph{k}-same-M algorithm \cite{gross2006model} based on Active Appearance Models (AAMs) \cite{matthews2004active} is proposed. This algorithm fits an AAM to input images then applies \emph{k}-same to the AAM model parameters. To further keep the data practically useful, Du \etal explicitly preserve race, gender and age attributes in face de-identification \cite{du2014garp}. Jourabloo \etal propose to jointly model face de-identification and attribute preservation in a unified optimization framework \cite{jourabloo2015attribute}.

The majority of sate-of-the-art approaches are designed based on the \emph{k}-same and implemented using AAM models. However, these methods have notable limitations: (\romannumeral1) The \emph{k}-same assumes that each subject is only represented once in the dataset, but this may be violated in practice. The presence of multiple images from same subject or images share similar biometric characteristics can lead to lower levels of privacy protection;  (\romannumeral2) \emph{k}-same operates on a closed set and produces a corresponding de-identified set, which is not applicable in situations that involve processing individual images or sequences of images; (\romannumeral3) Generated images by AAM models do not yet look natural enough.

Recently, GANs present promising effectiveness on image generating problems. It is a natural tool to synthesize de-identified images. Karla\etal \cite{brkic2017know} build a GAN-based model to generate full body images for de-identification, but the quality in face areas is not guaranteed. Bla{\v{z}} \etal use GAN to synthesize de-identified faces, but still based on the \emph{k}-same algorithm \cite{meden2017face}.

In this paper, we only focus on biometric information, i.e., we do not remove soft biometric features like age or race and non-biometric information like hairstyles \cite{ribaric2016identification}. We propose a novel \emph{Privacy-Protective-GAN} framework to address the problems mentioned above.

%-------------------------------------------------------------------------

\section{Method}
\label{sec:Method}
Face de-identification can be formulated as a transformation function $\delta$ which maps a given face image $x$ to a de-identified image $\hat{x}$, i.e. $\delta(x) = \hat{x}$, aiming to misleading the face verification. In this work, we propose a PP-GAN model and adopt the final generator as the de-identification function. First, we pretrain a verificator to determine whether two faces are from the same subject. Next, we freeze this verificator and utilize it to compute the contrastive loss in the whole system training phase. In each iteration, the synthesized output and the original image will be forwarded in: 1) the identity-related feature space by the pre-trained verificator to enforce the removal of identity; 2) the pixel level structure similarity by SSIM to preserve visual correspondence. Then the contrastive loss and the SSIM loss will be backpropagated to update the generator. Fig.~\ref{fig:framework} gives the big picture of the whole framework.

In the following subsections, we will describe the structure of the PP-GAN and explain its learning procedure in detail. The overview of the whole system is illustrated in Section \ref{sec:objective}. The detailed explanations of the Generator \emph{v.s.} Discriminator are given in Section \ref{sec:gan}. The verificator and structure similarity are introduced in  Sections \ref{sec:face_veri} and \ref{sec:struct_sim} respectively.

\subsection{Objective}
\label{sec:objective}
Face de-identification is a complex task involving multiple constraints at the same time. Specially, given an original face, we generate a new face image accordingly and hope that it meets three types of qualification: 1) The generated face should look natural and realistic, i.e. it looks like a face; 2) The generated face should not be discriminated as the same person with the original face, i.e. it should be far enough from the original one in the identity-related feature space; 3) The de-identified image should be consistent with the original image on the pixel level as much as possible, i.e., it shares similar luminance, contrast, and structure with the original image.

For the first objective, it can be achieved by utilizing the cGAN, which adversarially trains a generator ($G$) taking the original image as input and producing de-identified images versus the co-trained discriminator ($D$). The cGAN loss is denoted as $\mathcal{L}_{\mathrm{cGAN}}(G,D)$ here. The details of cGAN settings will be clarified in Section \ref{sec:gan}.

For the second goal, we pre-train a verificator to produce a verification loss in the identity embedding space for each pair of original and de-identified images. The verification loss for each generator $G$ is then formularized as the expected loss across subjects and denoted as $\mathcal{L}_{\mathrm{verif}}(G)$. This is used to guide the generator to enforce that the output holds an enough distance with the input in the identity-related feature space. The details about verificator will be elaborated in Section \ref{sec:face_veri}.

For the third objective, we adopt the Structural Similarity Index (SSIM) \cite{wang2004image} to quantify the similarity between the original and generated images. The SSIM loss combines the luminance, contrast, and structural differences in a product form with different power coefficients. Similarly as the verification loss, the SSIM loss for the generator $G$ is then formularized as the expected loss across subjects and denoted as $\mathcal{L}_{\mathrm{sim}}(G)$. The details of structure loss will be discussed in Section \ref{sec:struct_sim}.

Combining these three types of losses, our final objective of face de-identification is defined as
\begin{equation}
\label{eqn:final_obj}
\begin{split}
& \mathcal{L}_{\mathrm{face}} (G,D) \\
=& \mathcal{L}_{\mathrm{cGAN}}(G,D) + \lambda_1\mathcal{L}_{\mathrm{verif}}(G) +  \lambda_2\mathcal{L}_{\mathrm{sim}}(G),
\end{split}
\end{equation}
and the optimal generator $G^*$ is solved through the min-max procedure
\begin{equation} \label{eqn:final_obj_G}
G^* = \arg\min_G\,\max_D\,\mathcal{L}_{\mathrm{face}}(G,D),
\end{equation}
where $\lambda_1$ and $\lambda_2$ are the hyperparameters for multiple losses.

%-------------------------------------------------------------------------

\subsection{Architecture of GAN}
\label{sec:gan}
The cGAN learns a mapping $G:\{x,z\} \rightarrow \hat{x}$ from an observed image $x$ with additional random noise $z$ to a synthesized image $\hat{x}$, where $x$ is referred as a `real' sample or the condition from the original dataset, $\hat{x}$ is referred as a `fake' sample generated by the trained generator $G$, and $z$ is the random noise to ensure the image variability. The adversarial procedure trains a generator to produce outputs that can hardly be distinguished as a `fake' by the co-trained discriminator ($D$). Mathematically, the objective function of the cGAN is expressed as
\begin{equation}\label{eqn:cgan}
\begin{split}
\mathcal{L}_{\mathrm{cGAN}}(G,D) = \mathbb{E}_{x,\hat{x} \sim P_{\mathrm{data}}(x,\hat{x})}[\log D(x,\hat{x})]+\\
\mathbb{E}_{x \sim P_{\mathrm{data}}(x),z \sim P_{z}(z)}[\log(1-D(x,G(x,z))],
\end{split}
\end{equation}
where $G$ is the generator and $D$ is the discriminator. The optimal $G^*$ minimizes this objective against an adversarial $D$ that maximizes it, and it can be solved via a min-max procedure $G^* =	\arg\min_G\,\max_D\,\mathcal{L}_{\mathrm{cGAN}}(G,D)$. %Here in our task, the original probe face image is set as the input $x$ and the de-identified image $\hat{X}$ is then given by output $x$ from the generator $G$, i.e. $\hat{X} = x = G(x,z) = G(X,z)$, where $z$ is the random noise term.

\begin{figure*}[t]	
	\begin{center}
		\includegraphics[width=\linewidth,keepaspectratio]{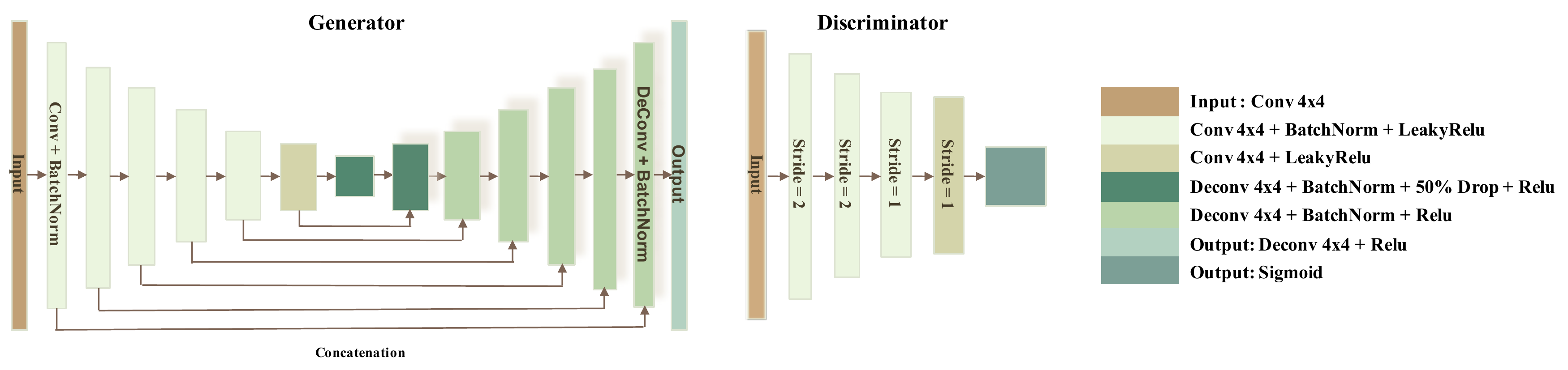}
	\end{center}
	\caption{
		\textbf {The network architecture of the generator G and the discriminator D.} The generator has a \enquote{U-net} architecture with input size of $1\times128\times128$, embedding size of $256\times1\times1$, and output size that is consistent with the input size.
		The input size of discriminator is $1\times128\times128$, while the output size is $1\times30\times30$.}
	\label{fig:GAN_arch}
\end{figure*}

In practice, we adapt our generator and discriminator architectures from the Image-to-Image translation \cite{isola2016image}. As shown in Fig.~\ref{fig:framework}, both of them use modules consisting of Convolution-BatchNorm-ReLu \cite{ioffe2015batch}. The generator is a ``U-Net'' \cite{ronneberger2015u} by adding skip connections between symmetric layers in the convolution and deconvolution steps. The stride keeps to be 2 so it resizes by 2 in the encoding part, expands by 2 in the decoding part. The size of the embedding layer is $1 \times 1$. Two drop-out layers in the middle serve as random noise $z$. For the discriminator, instead of feeding it with `real' or `fake' pairs, we follow the standard approach from \cite{goodfellow2014generative} and input either `real' image $x$ or `fake' image $\hat{x}$ since there is no unique solution for the de-identified $G(x)$. Also, a patch-design \cite{isola2016image} is adopted in the discriminator thus it outputs a $N \times N$ patch, rather than a single value to represent the probability of current input to be "real". Here we set N to 30 with receptive field of size 34. The network architectures are shown in Fig.~\ref{fig:GAN_arch}.

\subsection{Face Verification}
\label{sec:face_veri}
Face verification is to determine whether two images express the same person. Metric learning is a widely used approach for this task, which learns semantic distance measurements and the associated embeddings \cite{chopra2005learning,sun2014deep,schroff2015facenet,parkhi2015deep} from the original image space to feature space. Motivated by FaceNet \cite{schroff2015facenet}, we strive for an embedding $f(x)$ so that the squared distance between all faces are independent of imaging conditions.
In the embedding spaces, samples with the same identity are closer whereas those with different identities are further apart.

Mathematically, the embedding is denoted as $f: \mathrm{data}\rightarrow \mathbb{R}^d$, which embeds an image $x$ into a \emph{d}-dimensional Euclidean space. %We constrain this embedding to live on the \emph{d}-dimensional hypersphere, $\emph{i.e.} \left \| f(x) \right \|_2^{2} = 1$.
We adopt the following contrastive loss function originally proposed by Hadsell \etal \cite{hadsell2006dimensionality}. The loss function is defined as
\begin{equation}
\label{eqn:contrast_loss}
\begin{split}
& \mathrm{Verif}(x_i,x_j,\eta_{x_i,x_j},\alpha) = \\
&\left \{
\begin{aligned}
&\frac{1}{2} \left \|f(x_i)-f(x_j)\right \|_2^{2}, && \text{if}\ \eta_{x_i,x_j} = 0 \\
&\frac{1}{2}\max\left(0,\; \alpha -\left \| f(x_i)-f(x_j) \right \|_2\right)^{2}, && \text{if}\ \eta_{x_i,x_j}  = 1
\end{aligned} \right.
\end{split}
\end{equation}
where $\alpha$ is the margin that is enforced between positive and negative pairs, $f$ is the embedding function, and $\eta_{x_i,x_j}$ is the indicator where $\eta_{x_i,x_j} = 0$ means positive pairs, and $\eta_{x_i,x_j} = 1 $ means negative pairs.

For the network architecture, we choose the new Light CNN-9 model \cite{wu2015light}, which introduces the MaxFeature-Map (MFM) operation as an alternative of ReLU and claims that the MFM can play a role of feature selection. With this design, the Light CNN-9 model achieves state-of-the-art results with less parameters and time-consumption. The Light CNN-9 model contains 5 convolution layers, 4 Network in Network (NIN) layers \cite{lin2013network}, Max-Feature-Map function and 4 max-pooling layers. The final layer is a fully connected one that outputs a 256-dimensional representation. A detailed explanation of this architecture can be found in \cite{wu2015light}.

In the pre-training phase of the verificator, we randomly sample $50\%$ positive and $50\%$ negative pairs as train data to avoid data unbalance.
%with regard to the model balance.
These pairs are separately fed to a shared weight Siamese network, where the model is updated by the contrastive loss defined in Eq. (\ref{eqn:contrast_loss}). Once the verificator training is finished, parameters are frozen in the whole system training followed.

Next in the PP-GAN training phase, the pair of original image $x$ and synthesized image $\hat{x}$ are fed into pre-trained verificator with $\eta_{x,\hat{x}} = 1$ to enforce their different identities, i.e., the de-identification.

The term of verification loss in Eq. (\ref{eqn:final_obj}) is then defined as
\begin{equation}
\label{eqn:veri_loss}
\begin{split}
& \mathcal{L}_{\mathrm{verif}}(G)  \\
& =   \mathbb{E}_{x\sim P_{\mathrm{data}(x)},z \sim P_{z}} \left[\mathrm{Verif}(x,G(x,z),\eta,\alpha)\right],
\end{split}
\end{equation}
where $x$ is the original input $I$, $z$ is the noise, $G(x,z)$ is the generated image, $\eta = 1$ is the always false indicator, and the margin is set to $2$ on the normalized feature vector to enforce a positive loss.

%-------------------------------------------------------------------------
\subsection{Structure Similarity}
\label{sec:struct_sim}
For the de-identification problem, we want to hold the correspondence between original image $x$ and synthesized image $\hat{x}$. The vanilla GAN simply generates new samples from the data distribution. For cGAN, although it samples under the condition of the input image $x$, it does not ensure that the synthesized image $\hat{x}$ is close enough to $x$ in the image space. For example, if we have two images $x_i$ and $x_j$ with the same condition, the cGAN would generate $\hat{x_i}$ and $\hat{x_j}$ with the same distribution so that we cannot tell whether $\hat{x_i}$ is the de-identified image from $x_i$ or $x_j$. Thus an additional structural loss is necessary here to regulate the generator to assure this matching. The two popular choices, the mean squared error (MSE) and the related quantity of peak signal-to-noise ratio (PSNR), are not well suited because they do not match very well to the perceived visual quality by humans, and the pixel-level matching with the reference image is not what we want to optimize. Instead we use the \emph{Structural Similarity Index} (SSIM) \cite{wang2004image} as an objective measurement for assessing perceptual image degradation after de-identification. %This measure is proposed by Du \etal \cite{du2011preservative} to specially apply on the privacy protection problems .

SSIM consists of three components including luminance similarity $l(x, \hat{x})$, contrast similarity $c(x, \hat{x})$ and structural similarity $s(x, \hat{x})$:
\begin{equation}
\begin{split}
\mathrm{SSIM}(x, \hat{x}) = l(x, \hat{x})^{\alpha }\cdot c(x, \hat{x})^{\beta }\cdot s(x, \hat{x})^{\gamma }
\end{split}
\end{equation}
%where $x$ and $y$ are the reference image $I$ and de-identified image $\hat{I}$ respectively.
For a detailed explanation of SSIM, one can refer to \cite{wang2004image}.
Here we define $\mathcal{L}_{\mathrm{sim}}(x, \hat{x}) = \frac{1}{2}(1 - \mathrm{SSIM}(x, \hat{x}))$. Then in the GAN training, we have
\begin{equation}
\begin{split}
\mathcal{L}_{\mathrm{sim}}(G) =   \mathbb{E}_{x, \sim P_{\mathrm{data}(x)},z \sim P_{z}} \Big[\frac{1}{2}\big(1 - \mathrm{SSIM}(x, G(x,z)\big)\Big],
\end{split}\nonumber
\end{equation}
%where $x$ is the original input $I$, $z$ is the noise, and $G(x,z)$ is the generated image.
where $z$ is the noise, and $G(x,z)$ is the generated image.
%-------------------------------------------------------------------------
\section{Experiments}
\label{sec:Experiments}
\subsection{experimental Setup}

\begin{figure}[t]	
	\begin{center}
		\includegraphics[width=0.95\linewidth,keepaspectratio]{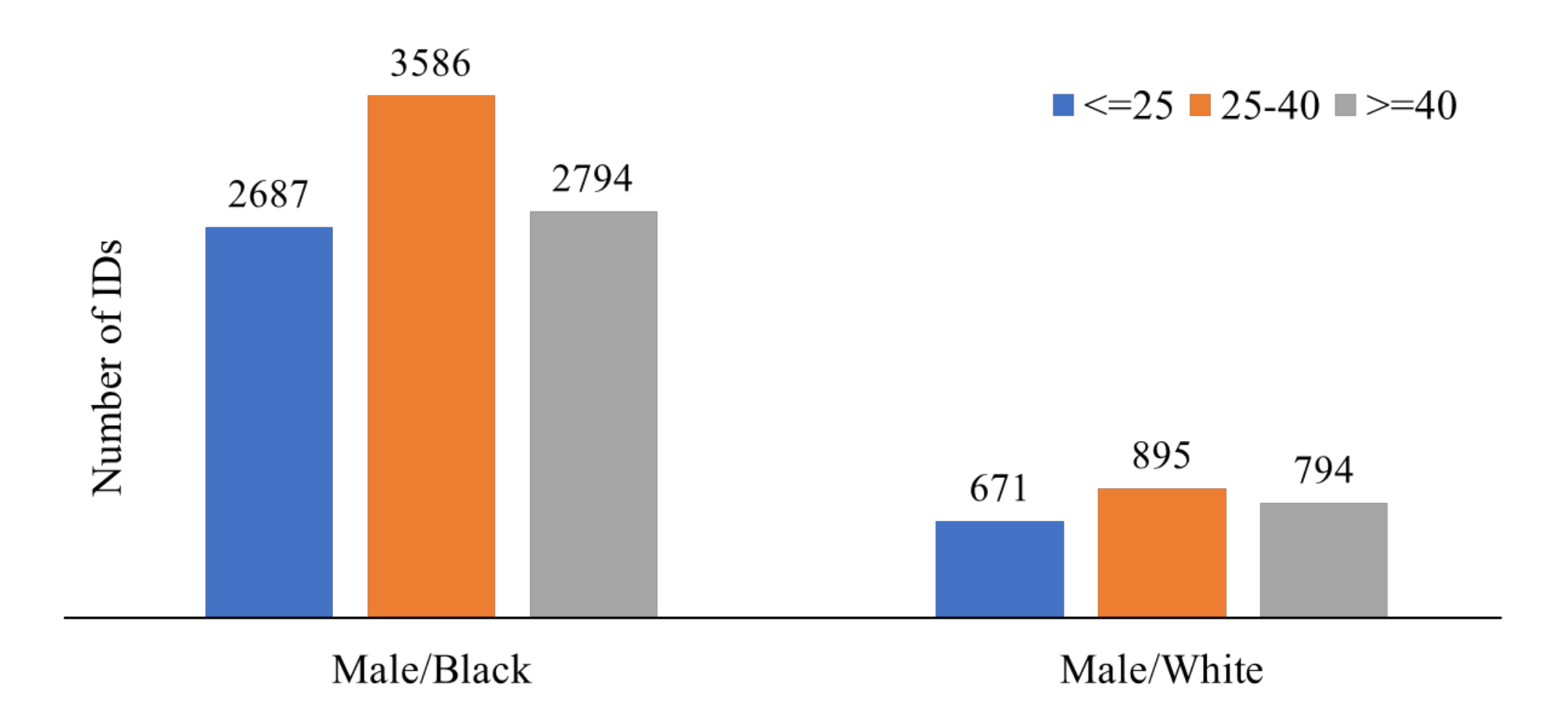}
	\end{center}
	\caption{\textbf{The demographic distribution of the MORPH dataset.} The black group contains more samples than the white group, while both have the similar age distribution.
	 }
	\label{fig:data}
\end{figure}

We evaluate the proposed framework on a publicly available MORPH \cite{ricanek2006morph} dataset, which contains 55,000 unique images of more than 13,000 individuals, with diverse demographic information like age, gender, and race. Here we only use male data since the number of femalel subjects is limited. The details of the demographic distribution in the MORPH dataset are shown in Fig.~\ref{fig:data}.

To validate that our framework can preserve attributes utility, data is divided into black and white according to race, then we divide each category into three age groups: youth (age $\leqslant$ 25), middle-aged (25 $<$ age $<$ 40) and senior (age $\geqslant$ 40). We refer these 8 different data groups as \emph{Black, White, Black-Youth, Black-Middle, Black-Senior, White-Youth, White-Middle} and \emph{White-Senior}. We randomly pick 90\% from each group for training and use the remaining for testing. The training and testing sets are consistent in the verificator pretraining and the whole system training. For the data preprocessing, we use OpenFace\footnote{\url https://cmusatyalab.github.io/openface} to detect, align, and crop face areas. During Verificator pretraining, we set initial learning rate as 1e-4 and margin as 2. During the whole system training, the learning rate is set to be 1e-5. For all the experiments, we use minibatch SGD and apply the Adam solver \cite{kingma2014adam}.

For each group, we separately conduct four experiments to analyze our objective function: only use cGAN, cGAN with SSIM loss, cGAN with pretrained verificator, amd cGAN with both pretrained verificator and SSIM constraint (i.e. ours). Here we refer these different set-ups as: \emph{cGAN-only,  cGAN+Sim, cGAN+Verif}, and \emph{cGAN+Sim+Verif}.

\subsection{Privacy Protection}
To demonstrate the proposed framework successfully generates face images for privacy protection, we need to prove that the produced face images neither remain the same as the corresponding raw images, which means successful de-identification, nor simply changing the identity to other identities in dataset, which means no switching identity occurs.

\textbf{De-identification}
To evaluate if the proposed model successfully removes the identity of original images, we assess if a pair of images are from same identity in a verification way. Specifically, given a verificator, for each pair of images $(x,\hat{x})$, it gives a distance value indicating the similarity of the two images. If the value is smaller than a threshold, the two images are treated as the same identity, which means de-identification fails; otherwise, from different identities, which means de-identification successes. Here we use de-identification rate (ERR rate) to represent the effectiveness of privacy protection, i.e., $(x,\hat{x})$ is not determined as the same person.

To eliminate the overfitting caused by the pre-trained verificator since it guides the training of the whole system, we fine-tune the FaceNet \cite{schroff2015facenet}\footnote{\url https://github.com/davidsandberg/facenet} with triplet loss as our evaluation verificator.  The FaceNet is trained on the 8 subgroups respectively. Specifically, the test data are grouped into numerous pairs with half positive and half negative. Then the trained model is evaluated on these pairs of images in a 10-fold way. The finally gained optimal threshold is utilized in de-identification evaluation.

\begin{table}[]
\caption{De-identification rate (percent) of Male/Black.}	
	\centering
	\small
	\resizebox{\linewidth}{!}{%
		\begin{tabular}{c|cccc}
			\hline\hline
			& Black& Black-Youth& Black-Middle& Black-Senior \\ 	\hline
			Original Test & 1.5 & 1.6& 1.6& 2.7\\ 		
			De-identified Train  & 100.0 & 100.0       & 100.0       & 97.2       \\
			De-identified Test   & 100.0 & 100.0       & 97.0         & 93.7     \\ 	\hline
		\end{tabular}
	}

	\label{table:De-id-Black}
\end{table}

\begin{table}[]
\caption{De-identification rate (percent)  of Male/White.}	
	\centering
	\small
	
	\resizebox{\linewidth}{!}{%
		\begin{tabular}{c|cccc}
			\hline\hline
			& White & White-Youth& White-Middle & White-Senior \\\hline
			Original Test & 1.8 & 2.8& 3.3& 5.9\\ 		
			De-identified Train  & 100.0 & 96.2       & 89.7        & 91.7       \\
			De-identified Test  & 100.0 & 94.4      & 90.8        &84.7      \\ \hline
		\end{tabular}
	}
    \label{table:De-id-White}
\end{table}

Table \ref{table:De-id-Black} and Table \ref{table:De-id-White} show our de-identification performance. The FaceNet achieves accuracy greater than 97\% on all Black groups and 94\% on White groups. The overall de-identification rates are both very high, and that of the black group is a bit higher due to the larger sample size.

The majority state-of-the-art de-identification algorithms based on k-anonymity theory, and k-value will directly affect the de-identification rate. The APFD algorithm \cite{jourabloo2015attribute} shows the performance of de-identification saturates when $k = 8$, and achieves the de-identification rate around 90\%, note that the Bayesian classifier used for evaluation only achieves verification accuracy of 70\%. Recently GAN-based face de-identification achieves best de-identification rate of 43.2\% when $k=2$ \cite{meden2017face}.
Thus it is obviously that our system outperforms previous methods regrading the effectiveness of privacy protection by explicitly adding de-identification metric into the optimization objective.

\textbf{No switching-identification} In addition, in order to demonstrate the yielded image does not possess the identity of the other original images, for each generated image, we compute its similarity, i.e., distance, with every original image, and compare the distance of image pair $(x_i,\hat{x_j})$ with a threshold, where $i,j$ are the identity of image. If the distance is smaller than the threshold and $i\neq j$, identification switching occurs. If the distance is smaller than the threshold and $i= j$, identification switching does not occur. We use the number of identification switches (IDS) times as a measurement.

We conduct experiments on \emph{Black-Youth} and \emph{White-Youth}.
As the data are unbalanced, to fairly compare the IDS on both groups, we randomly sample same number of people from \emph{Black-Youth} test as that in \emph{White-Youth} test. On both groups, we obtain 0 IDS, which means no identification switching occurs.

\subsection{Utility Preservation}

\textbf {Face-like}
The most important point for data utility is that the generated images should look natural and realistic, i.e, they look like face. Here we utilize a face detector
MTCNN \cite{mtcnn} to determine whether the generated images are face or not. To alleviate the variance of the face detector, we conduct experiments on both original and generated face images. Table~\ref{table:Detect-Black} and \ref{table:Detect-White} show the detection rate on both the original and generated data. From the tables, we note that the detection rates on original and de-identified data are similar, proving that our system preserves the face-like utility very well. As cropped face images are not suitable for face detector trained on face image with context, to alleviate the effects from the inconsistency, we pad the cropped image with 50 pixels of 0 along each axis. For some groups, the detection rate on de-identified images is higher than that on original images, e.g. \emph{Black-Youth}. This is because some original face images are distorted, but yielded images have less distorted images.
%\emph{Black, White, Black-Youth, Black-Middle, Black-Senior, White-Youth, White-Middle} and \emph{White-Senior}.
\begin{table}[t]
	\caption{Detection rate of Male/Black.}	
	\centering
	\small
	\resizebox{\linewidth}{!}{%
%		\begin{tabular}{c|cccc}
		\begin{tabular}{@{\hspace{0.mm}}c@{\hspace{0.5mm}} | @{\hspace{0.75mm}}c@{\hspace{0.75mm}} @{\hspace{0.75mm}}c@{\hspace{0.75mm}}@{\hspace{0.75mm}}c@{\hspace{0.75mm}}@{\hspace{0.75mm}}c@{\hspace{0.mm}}}
			\hline\hline
			Data Settings& Black& Black-Youth& Black-Middle& Black-Senior \\
			\hline
			Original w/o padding & 0.733 & 0.817 & 0.767 & 0.598\\
			De-identified w/o padding& 0.719 & 0.847 & 0.715 & 0.570\\
			Original w padding & 0.988 & 0.993 & 0.994 & 0.977\\
			De-identified w padding& 0.956 & 0.995 & 0.990 & 0.969\\
			\hline			
		\end{tabular}
	}
	
	\label{table:Detect-Black}
\end{table}
\begin{table}[]
	\centering
	\small
	\caption{Detection rate of Male/White.}	
	\resizebox{\linewidth}{!}{%
%		\begin{tabular}{c|cccc}
		\begin{tabular}{@{\hspace{0.mm}}c@{\hspace{0.5mm}} | @{\hspace{0.75mm}}c@{\hspace{0.75mm}} @{\hspace{0.75mm}}c@{\hspace{0.75mm}}@{\hspace{0.75mm}}c@{\hspace{0.75mm}}@{\hspace{0.75mm}}c@{\hspace{0.mm}}}
			\hline\hline
			Data Settings& White & White-Youth& White-Middle & White-Senior \\
			\hline
			Original w/o padding & 0.758 & 0.808 & 0.738 & 0.621\\
			De-identified w/o padding& 0.784 & 0.767 & 0.739 & 0.499\\
			Original w padding &0.973 & 0.967 & 0.979 & 0.975\\
			De-identified w padding&0.975& 0.940 & 0.988 & 0.902\\
			\hline			
		\end{tabular}
	}
	\label{table:Detect-White}
\end{table}

\begin{figure}[]	
	\begin{center}
		\includegraphics[width=0.8\linewidth,keepaspectratio]{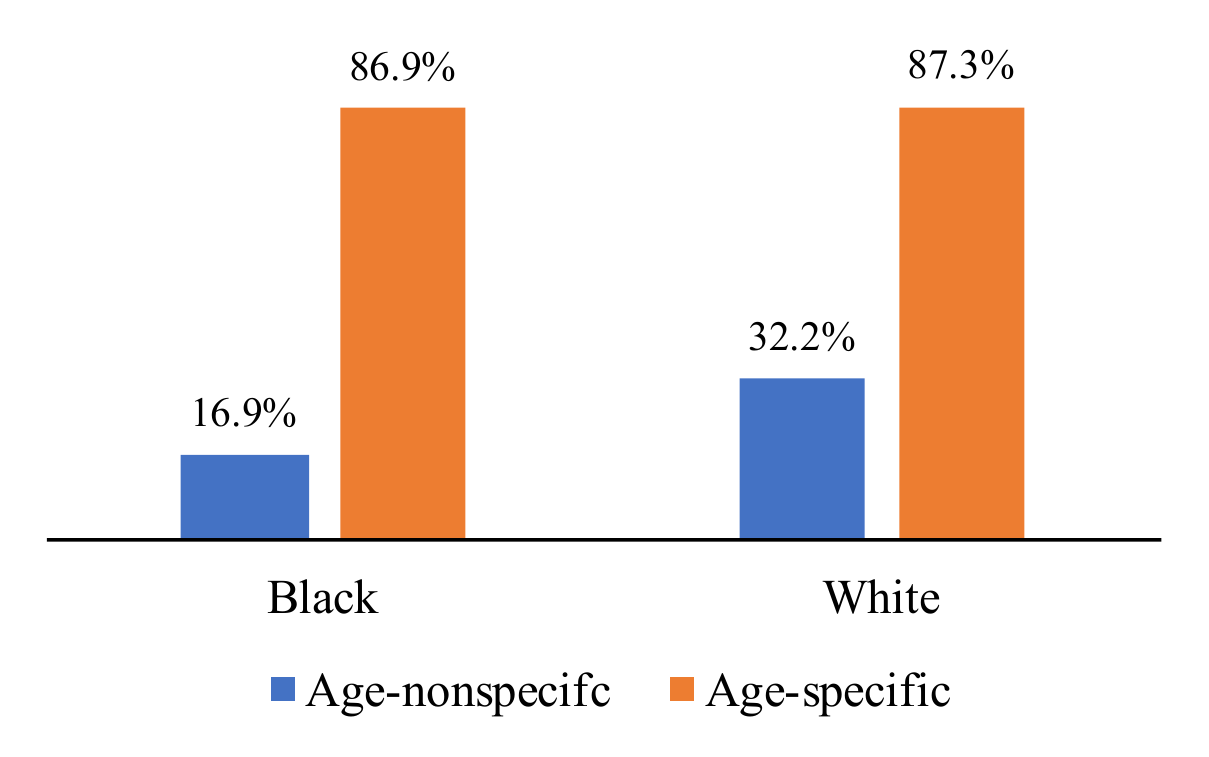}
	\end{center}
	\caption{\textbf{Age classification accuracy.}  The blue bar displays the classification accuracy of preserving attributes for age-nonspecific group and the orange bar is for age-specific group.
	}
	\label{fig:age}
\end{figure}

\textbf {Attributes Preservation}
To examine the performance of the proposed model for attribute preservation, we train separate classifiers for each attribute and compute the classification accuracy for the generated images. The classification accuracy rate is used as a measurement to evaluate experimental results. From Figure~\ref{fig:age}, we can see that the age-specific models well preserve its group attributes compared to the age-nonspecific group. The experiments on original images display similar comparisons.

\subsection{Visual Similarity}
\begin{figure}[t]	
	\begin{center}
		\includegraphics[width=\linewidth,keepaspectratio]{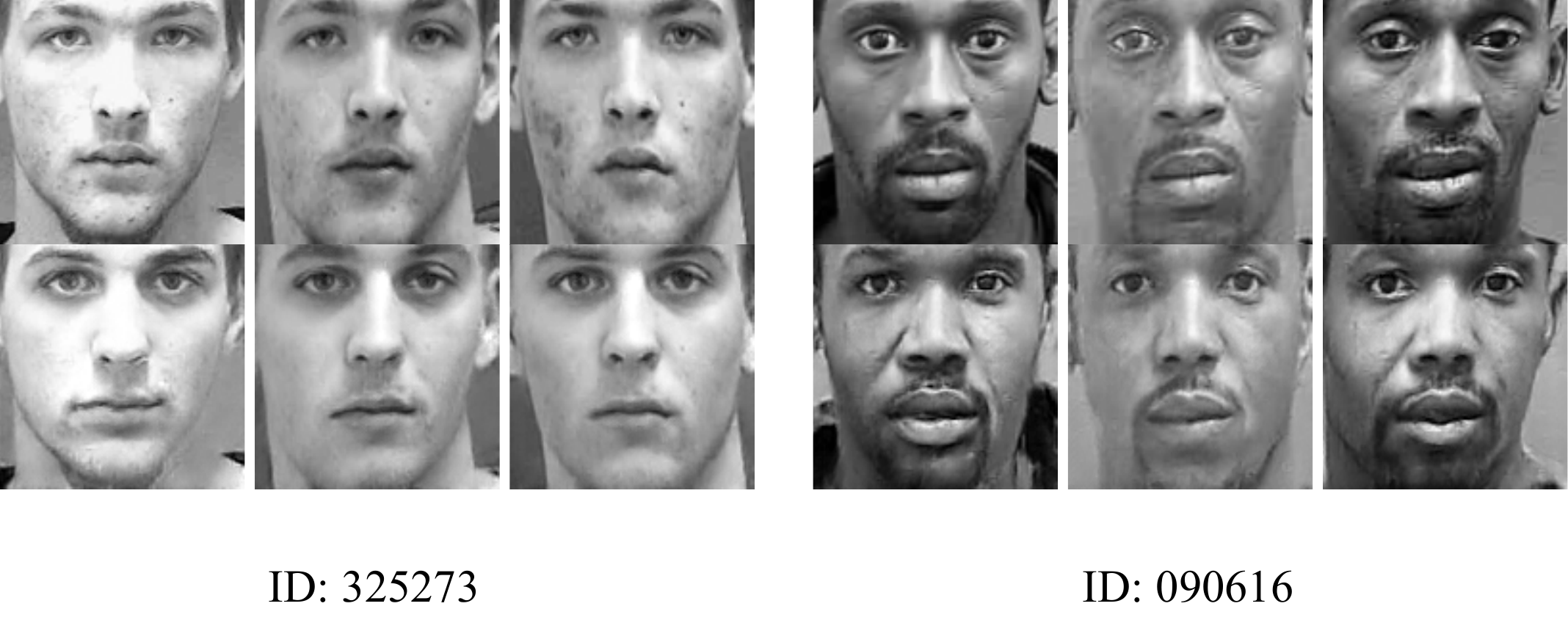}
	\end{center}
	\caption{\textbf{The visualization of correspondence between original and de-identified images.}  The top row shows the original face images, the bottom row shows the corresponding de-identified images synthesized by our PP-GAN model.
	}
	\label{fig:corr}
\end{figure}

For the de-identification problem, in addition to the utility preservation, we want to hold the correspondence between original image $x$ and synthesized image $\hat{x}$ rather than just replace the face area randomly. In other words, we only want to remove the privacy related characteristics, but keep the visual similarity, e.g. contours and luminous condition, as much as possible. As shown in Fig.~\ref{fig:corr}, the original and generated images share consistent SSIM-consisting features like luminance, contrast, and structure information while distinguish in  privacy related characteristics such as the sizes and shapes of eyes, nose, and mouth. This indicates the effectiveness of the regulator.
%In this way, the data utility can be preserved to the utmost extent for various applications. the %For instance, generated images can be used to replace face area in the surveillance videos smoothly.

\begin{figure*}[t]	
	\begin{center}
		\includegraphics[width=\linewidth,height=.17\linewidth]{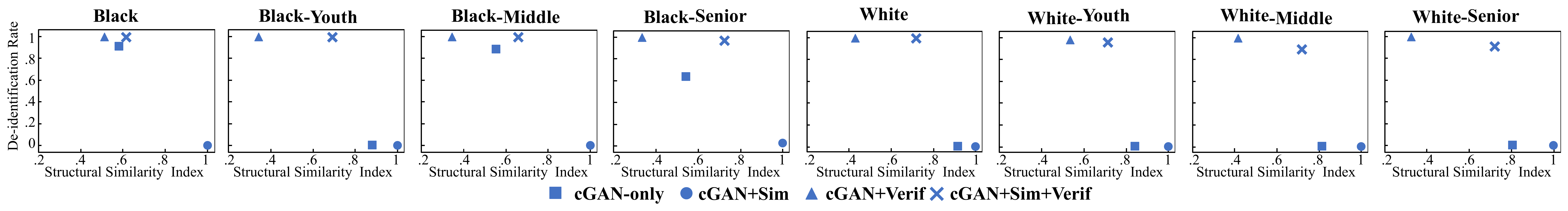}
	\end{center}
	\caption{\textbf{The trade-off between identity removal and visual similarity.} The horizontal axis represents values of Structural Similarity Index (SSIM), the vertical axis represents the de-identification rate.
	}
	\label{fig:ssim}
\end{figure*}

\begin{figure*}[t]	
	\begin{center}
		\includegraphics[width=\linewidth,keepaspectratio]{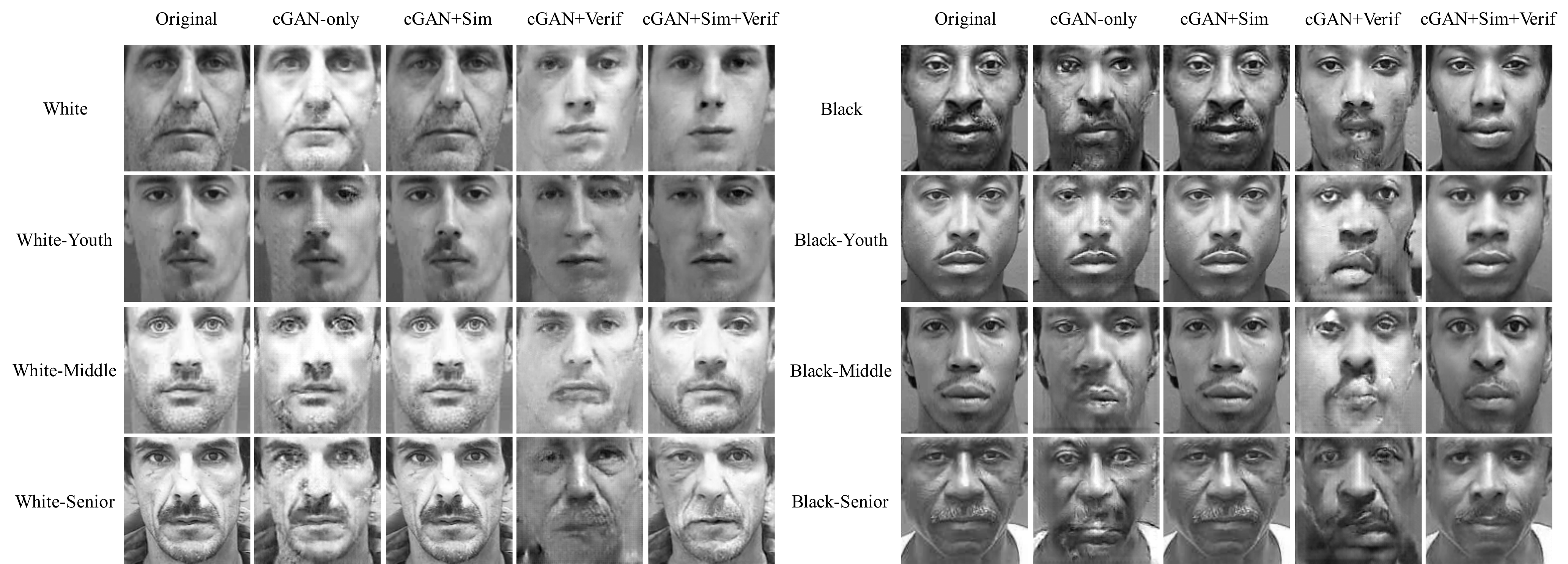}
	\end{center}
	\caption{\textbf{The visualized results of different objective designs on different groups.} Our PP-GAN generates visually agreeable results beyond all the other settings on every group.
	}
	\label{fig:re-vis}
\end{figure*}

%\begin{figure}[t]	
%	\begin{center}
%		\includegraphics[width=\linewidth,keepaspectratio]{fail.png}
%	\end{center}
%	\caption{Fail examples
%	}
%	\label{fig:fail}
%\end{figure}

To demonstrate the proposed framework achieves both de-identification and visual similarity, we compute id-removal rate and SSIM of each yielded image with the corresponding original image on \emph{cGAN-only, cGAN + Verif, cGAN + Sim}, and \emph{cGAN + Verif + Sim} (see Fig.~\ref{fig:ssim}), and display representative results (see Fig.~\ref{fig:re-vis}). Overall, our framework maintains high id-removal rate and SSIM at the same time. One can also visually tell that the PP-GAN model maps the original face to a different face while somewhat keeps certain implicit relationship.

For the \emph{cGAN-only} setting, it shows a high id-removal rate with uncertain SSIM for the black (Fig.~\ref{fig:ssim} \emph{Left-half}) and a low id-removal rate with high SSIM for the white (Fig.~\ref{fig:ssim} \emph{Right-half}), which may come from the different sizes of samples. When the sample size is small, the cGAN-only generates new images quite similar to existing one (Fig.~\ref{fig:re-vis} \emph{Left}) and gains high SSIM. When the sample size is large, the cGAN-only generator is able to step away from the original figures but may generate improper visualizations (Fig.~\ref{fig:re-vis} \emph{Right}) and provides low insurance for SSIM.

For the \emph{cGAN+Sim} setting, the SSIM cost degenerates cGAN to the same manner of cGAN with limited samples, which functions almost as an identity mapping (Fig.~\ref{fig:re-vis}). Thus it holds low id-removal rate and high SSIM.

For the \emph{cGAN+Verif} setting, the verification loss here enlarges the cGAN's sampling range thus results in a high id-removal rate (Fig.~\ref{fig:ssim}), while the inconsistency between the two feature spaces and lack of structure constrains may cause the visually strange consequences (Fig.~\ref{fig:re-vis} \emph{Right}). An interesting phenomenon is that when the sample size is small (the white group), the \emph{cGAN+Verif} indeed obtains a visually moderate result (Fig.~\ref{fig:re-vis} \emph{left}) due to the enlarged sampling range from the verificator.

These results confirm the necessity and efficiency of simultaneously including the verificator and regulator to balance the id-removal and utility retaining.

%-------------------------------------------------------------------------
\section{Conclusion}
\label{sec:conlcusion}
In this paper, we present a new face de-identification framework, Privacy-Protective-GAN, which generates de-identified output according to a single input. We explicitly integrate the de-identification metric into the objective function to ensure the privacy protection. Meanwhile, we try to preserve visual similarity as much as possible to retain data utility by adding a regulator.
In the experiments, we quantitatively demonstrate the effectiveness of proposed method in terms of privacy protection, utility preservation, and visual similarity.

%We can see that on Black, Black-Middle, and Black Senior dataset SSIM rates of \emph{cGAN-only} are inferior and de-identification rates of \emph{cGAN-only} are superior.

%The same phenomenon also happens in results of \emph{cGAN+ Verif} This is because the \emph{cGAN-only} and \emph{cGAN+ Verif} fail to produce natural and realistic face images. Figure {?} shows...

%On Black-Youth, White, White-Youth, White-Middle, and White-Senior, \emph{cGAN-only, cGAN+ Sim} have low de-identification rates and SSIM. The reason behind the phenomenon is \emph{cGAN-only, cGAN+ Sim} generate natural and realistic images but these images are same with raw images. Figure{?} shows...
%On all dataset, \emph{cGAN+ Sim} has high SSIM and low de-identification rates. This means \emph{cGAN+ Sim} yielded images are similar or identical with original images. Figure{?} shows that ....

%\emph{cGAN + Verif + Sim} gains high de-identification and SSIM values. This means that generated images are natural, realistic, and de-identified images. Figure {?}....
%Then, we average the SSIMs over yielded images to obtain the mean of SSIM on a dataset. The mean value is used as a measurement to compare four different set-ups, and illustrate the influence of the SSIM loss.

%-------------------------------------------------------------------------
\clearpage
{\small
\bibliographystyle{ieee}
\bibliography{ppgan}
}

\begin{figure*}[h]	
	\begin{center}
		\includegraphics[width=0.90\linewidth,keepaspectratio]{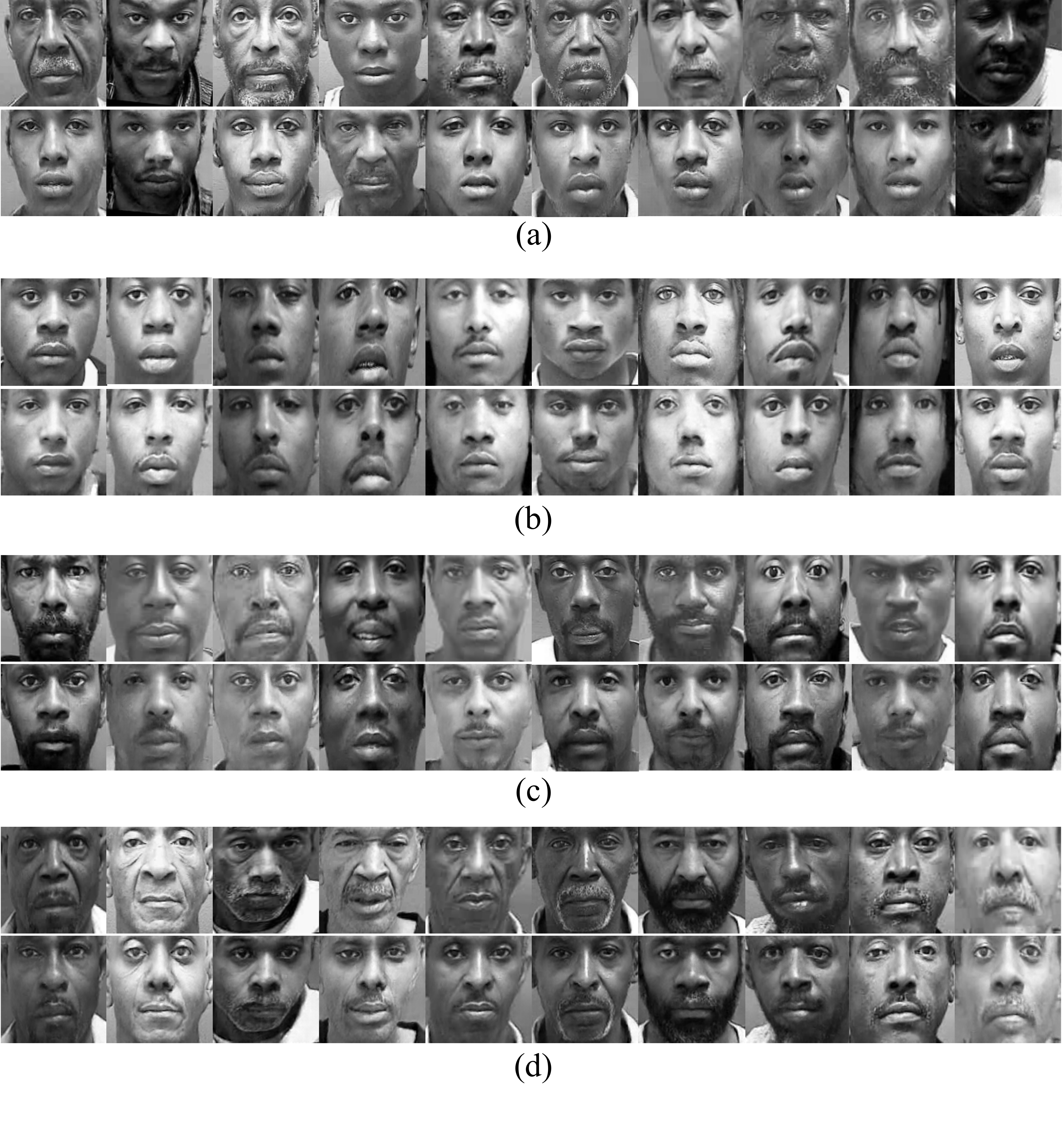}
	\end{center}
	\caption{\textbf{The visualized results of the black group.} In each pair of double rows, the top one is for the original faces and the bottom one is for the de-identified faces. \emph{(a)} This panel shows the results for the whole black group. We can see that the age factor is not retained. The \emph{(b)} shows the results of \emph{Black-Youth} subgroup. The \emph{(c)} shows the results of \emph{Black-Middle} subgroup. The \emph{(d)} shows the results of {Black-Senior} subgroup.
	}
\end{figure*}

\begin{figure*}[h]	
	\begin{center}
		\includegraphics[width=0.90\linewidth,keepaspectratio]{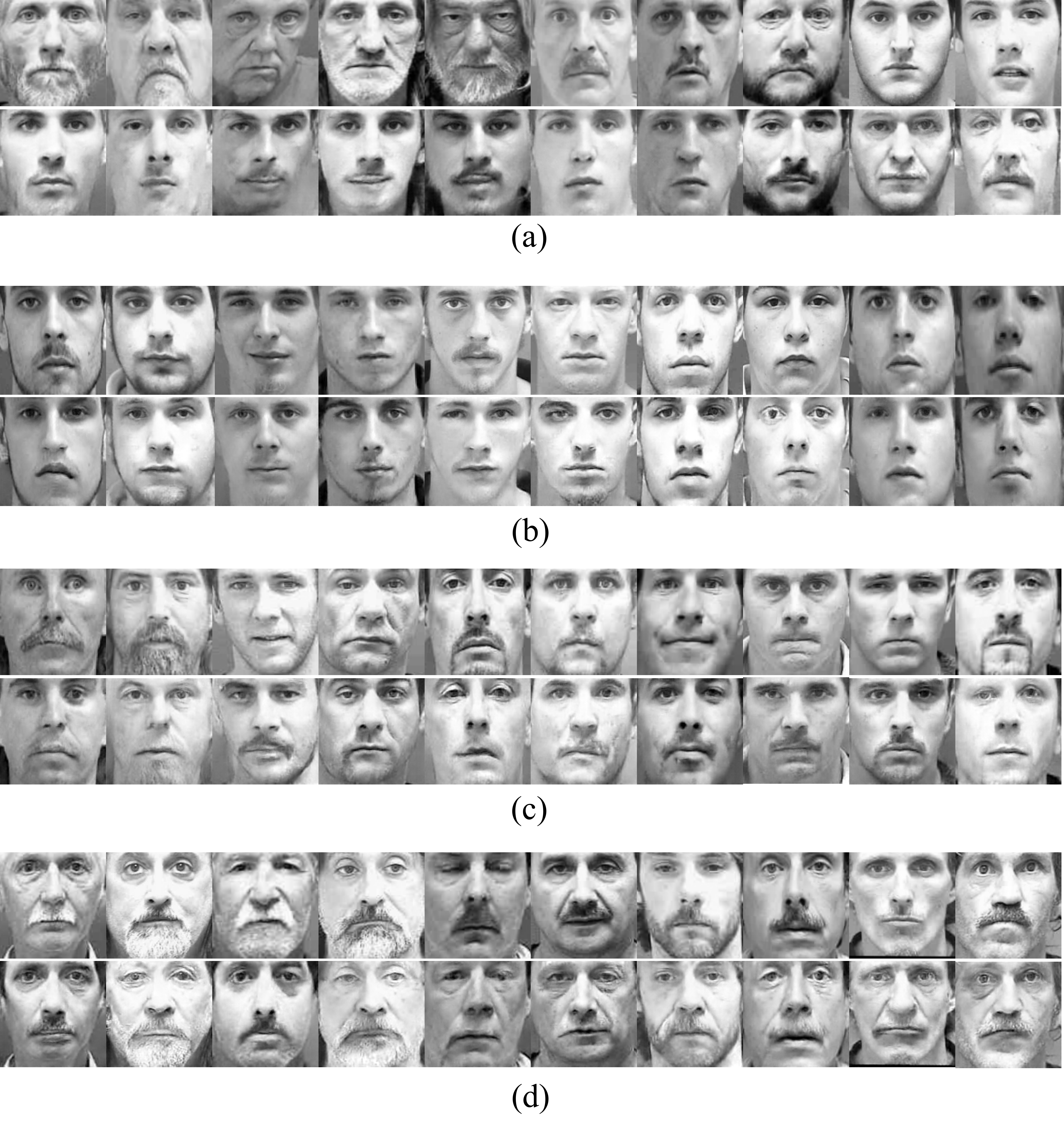}
	\end{center}
	\caption{\textbf{The visualized results of the white group.} In each pair of double rows, the top one is for the original faces and the bottom one is for the de-identified faces. \emph{(a)} This panel shows the results for the whole white group. We can see that the age factor is not retained. The \emph{(b)} shows the results of \emph{White-Youth} subgroup. The \emph{(c)} shows the results of \emph{White-Middle} subgroup. The \emph{(d)} shows the results of {White-Senior} subgroup.
	}
\end{figure*}
\end{document}